\documentclass[final,3p,times,authoryear]{elsarticle}

\usepackage{comment}
\usepackage{booktabs}
\usepackage{amssymb}
\usepackage{amsmath}
\usepackage{tipa}

\journal{Computer Speech \& Language}

\begin{document}

\begin{frontmatter}

\title{Assessing the Ability of Neural TTS Systems to Model Consonant-Induced F0 Perturbation}

\author[inst1]{Tianle Yang\corref{cor1}}
\author[inst2]{Chengzhe Sun}
\author[inst3]{Phil Rose}
\author[inst1]{Cassandra L. Jacobs}
\author[inst2]{Siwei Lyu}

\cortext[cor1]{Corresponding author: tianleya@buffalo.edu}

\affiliation[inst1]{
    organization={University at Buffalo},
    addressline={Department of Linguistics}, 
    city={Buffalo},
    postcode={14260},
    state={NY},
    country={United States}
}

\affiliation[inst2]{
    organization={University at Buffalo},
    addressline={Department of Computer Science and Engineering}, 
    city={Buffalo},
    postcode={14260}, 
    state={NY},
    country={United States}
}

\affiliation[inst3]{
    organization={Australian National University},
    addressline={Emeritus Faculty},
    city={Canberra},
    postcode={0200},
    state={ACT},
    country={Australia}
}

\begin{abstract}
This study proposes a segmental-level prosodic probing framework to evaluate neural TTS models’ ability to reproduce consonant-induced f0 perturbation, a fine-grained segmental-prosodic effect that reflects local articulatory mechanisms. We compare synthetic and natural speech realizations for thousands of words, stratified by lexical frequency, using Tacotron 2 and FastSpeech 2 trained on the same speech corpus (LJ Speech). These controlled analyses are then complemented by a large-scale evaluation spanning multiple advanced TTS systems. Results show accurate reproduction for high-frequency words but poor generalization to low-frequency items, suggesting that the examined TTS architectures rely more on lexical-level memorization rather than abstract segmental-prosodic encoding. This finding highlights a limitation in such TTS systems' ability to generalize prosodic detail beyond seen data. The proposed probe offers a linguistically informed diagnostic framework that may inform future TTS evaluation methods, and has implications for interpretability and authenticity assessment in synthetic speech.

\end{abstract}


\end{frontmatter}

\section{Introduction}
In recent years, neural text-to-speech (TTS) systems have achieved remarkable naturalness, enabled by several advances including end-to-end modeling \citep{wang2017tacotron, kim2021conditional}, improved prosody prediction \citep{ren2020fastspeech, ye2023clapspeech}, and high-fidelity neural vocoders \citep{van2016wavenet, kong2020hifi}. While these improvements have led to more fluent and expressive synthetic speech, current evaluations have primarily focused on higher-level prosodic features. However, the ability of TTS models to replicate fine-grained phonetic effects remains underexplored. One such effect is consonant-induced $f_{0}$ perturbation (hereafter referred to as $f_{0}$ perturbation), a well-documented phenomenon in phonetics where the fundamental frequency ($f_{0}$) of a vowel is systematically influenced by the immediately preceding consonant \citep{kirby2016effects, xu2021consonantal, gao2024laryngeal, yang2025onset}. Unlike high-level prosodic patterns, $f_{0}$ perturbation effects often arise from short-range dependencies between local segmental context, such as the voicing, aspiration, and glottal tension of a preceding consonant that are not explicitly supervised during the training of TTS models. If such perturbation effects were to emerge in synthetic speech, this would suggest that a TTS model has internalized not just surface-level statistical associations, but also the systematic co-variation between segmental features and continuous acoustic parameters. This makes perturbation an ideal linguistically grounded probe for evaluating how deeply such systems encode structured articulatory-acoustic relationships.

\subsection{Background}
\subsubsection{Previous work on F0 perturbation}
To better understand what it means for a TTS system to reproduce $f_0$ perturbation, we briefly summarize the patterns observed in natural speech. The most widely observed consonant- induced $f_0$ perturbation is that voiceless obstruents tend to induce a higher $f_{0}$ on the following vowel than their voiced counterparts. While the effect of voiceless obstruents is robustly attested across languages, the effect of voiced obstruents varies: some studies report a lowering of $f_{0}$ \citep{chen2011does, rose2022modelling}, others find a neutral effect \citep{hanson2009effects, yang2025onset}, depending on the local and global prosody \citep{xu2003effects}, annotation and normalization method \citep{xu2021consonantal}, or language specific phonetics and phonology \citep{zhang2020effect}. These perturbation patterns are generally attributed to well-understood physiological mechanisms. The higher $f_{0}$ after voiceless obstruents is typically attributed to increased vocal fold tension \citep{park2019effects}. This gesture suppresses voicing during oral closure by making the vocal folds stiffer and less able to oscillate, especially in the absence of a strong transglottal pressure gradient \citep{halle1971note, kirby2016effects}. In contrast, the lower $f_{0}$ associated with voiced obstruents is linked to laryngeal lowering, which facilitates sustained voicing during closure \citep{hoole2011automaticity, sole2018articulatory}.

In addition to the $f_{0}$ perturbation effect induced by the preceding consonants, the $f_{0}$ realization of a vowel is also influenced by the vowel itself. High vowels such as [i] tend to be produced with a higher $f_{0}$ compared to low vowels such as [\textipa{6}] \citep{van2011intrinsic, jacewicz2015intrinsic}. This phenomenon, often referred to as the vowel
intrinsic F0 (VF0) effect, is observed across a wide range of languages and is also thought to reflect biomechanical differences in the production of high versus low vowels \citep{whalen1995universality, whalen1995intrinsic}. Although the VF0 effect is not the focus of our study, it is nonetheless important to carefully model this effect so that any observed $f_0$ differences can be more reliably attributed to the effects of consonantal context rather than to vowel properties. In the present experiment, we controlled for vowel height by categorizing vowels according to the most recent IPA classification, grouping them into three levels: high, mid, and low. Other factors, such as the aspiration of a consonant, may also influence $f_{0}$ perturbation, but prior research suggests that these effects are relatively subtle and short-lived \citep{lofqvist1989cricothyroid, clements2007phonetic, schertz2020acoustic} compared with the CF0 and VF0 effects described above.

\subsubsection{Comparison Baseline for F0 perturbation}
Because the consonant-induced $f_{0}$ perturbation is by definition a relative effect, a suitable baseline is necessary to quantify and interpret any pitch differences following different consonant types. In this study, following the approach originally proposed by \cite{hanson2009effects} and subsequently validated by \cite{kirby2016effects}, we adopt vowels following sonorant consonants (such as [\textipa{m}], [\textipa{n}], [\textipa{N}]) as our reference or baseline condition.

The choice of sonorants as a reference rests primarily on physiological grounds. Sonorants such as the nasal consonant involve minimal adjustments in the supralaryngeal cavity or the state of the glottis to maintain the vibration of the vocal fold \citep{hanson2009effects}. Thus, sonorants provide an articulatorily neutral point against which perturbations induced by other consonant types can be reliably measured. In addition, previous evidence from \cite{kirby2016effects} further justifies the selection of sonorant-initial vowels as a baseline. They demonstrated that vowels following sonorants, especially nasals, show consistent and stable $f_{0}$ trajectories across different languages. This pattern further shows the reliability of sonorant consonants as a stable reference for quantifying onset-related perturbation.

\subsection{Comparing different model structure}
Given the automatic nature of these acoustic results\footnote{Note that the physiological mechanisms itself is not completely automatic, but may instead reflect some degree of speaker-level control or intentional modulation \citep{kingston1994phonetic, zhang2020effect}. From a control-based perspective, the magnitude and duration of consonant-induced $f_{0}$ perturbation (CF0) and vowel-induced $f_{0}$ perturbation (VF0) effects may vary across languages depending on the functional need to enhance particular phonological contrasts. For instance, enhancing voicing through stronger CF0 cues, or reinforcing vowel height via more pronounced VF0 patterns.}, their presence in synthetic speech may depend on how sensitively the model encodes fine-grained segmental interaction. This consideration becomes especially relevant when comparing different TTS architectures. Autoregressive models such as Tacotron 2 \citep{shen2018natural} generate acoustic frames sequentially, with each frame conditioned on the previous ones. This frame-level temporal recursion allows acoustic dependencies to be implicitly propagated and shaped over time. In contrast, non-autoregressive models such as FastSpeech 2 \citep{ren2020fastspeech} generate acoustic representations in parallel and therefore lack such feedback. To compensate for this architectural constraint, temporal structure is imposed through predicted durations and positional encodings, while the $f_{0}$ contour is modeled explicitly via a dedicated pitch predictor rather than emerging implicitly through sequential decoding. These architectural differences imply distinct mechanisms by which segmental effects, such as onset-related $f_{0}$ perturbation, may be learned and generalized.

To investigate whether TTS architecture influences the realization of segmental $f_{0}$ effects, we focus on two representative models: Tacotron 2 and FastSpeech 2. These two models represent distinct generation paradigms: Tacotron 2 uses an autoregressive mechanism that produces each frame conditioned on previous outputs, whereas FastSpeech 2 generates all frames in parallel. This contrast allows us to examine whether generation strategy affects the model’s ability to reproduce segmentally conditioned $f_{0}$ perturbation patterns. Another motivation for selecting these models is the availability of a consistent baseline. Both models are trained on the LJSpeech corpus \citep{ljspeech17}, a single-speaker dataset that provides the natural recordings (human speech) used as our ground-truth reference. This setup allows for a direct comparison between synthetic and natural realizations of the same items from the same speaker. Lastly, we also considered the accessibility of both models. Tacotron 2 and FastSpeech 2 offer open training data, pretrained weights, and clear architectural documentation, which makes the results easily reproducible and interpretable. This level of accessibility contrasts with most TTS systems, which are often partially open-access or entirely proprietary.

\subsection{Implication}
In sum, this study presents the first systematic evaluation of whether neural TTS models capture onset-induced $f_{0}$ perturbation, a segmental-level phonetic effect. We also examine whether differences in model architecture, particularly the distinction between autoregressive and non-autoregressive generation, influence this ability. Our results show that neither Tacotron 2 nor FastSpeech 2 accurately captures these effects in low-frequency or unseen items. This finding is later confirmed by a large-scale study in section \ref{sec:generalize}, which suggests that TTS architectures may benefit from mechanisms that more explicitly model segmental-prosodic interactions.

\section{Experimental Setting}
We conduct two sets of experiments in this study. Experiment 1 (section \ref{sec:Exp1}) offers a theory-driven evaluation of whether neural TTS models (Tacotron 2 and FastSpeech 2) can reproduce segmental $f_0$ perturbation patterns, by directly comparing synthetic and natural speech from the same single-speaker corpus (LJ Speech), while Experiment 2 (section \ref{sec:generalize}) demonstrates the practical transferability of these phonetic cues in a large-scale, multi-speaker setting, using real and fake speech to assess their robustness and diagnostic utility.

\subsection{Speech generating and alignment}
In addition to the pre-segmented LJ Speech dataset described above, we synthesized 4,210 sentences using FastSpeech 2 and Tacotron 2, both trained exclusively on the LJ Speech corpus. The sentences were randomly sampled from the Corpus of Contemporary American English (COCA), a balanced corpus comprising approximately one billion words drawn from a range of genres, including spoken language, fiction, magazines, newspapers, and academic texts, spanning from 1990 to 2015 \citep{davies_coca_hd}. The generated audio was aligned with its corresponding text using the MFA (Montreal Forced Aligner; \citealp{mcauliffe2017montreal}), which produced phoneme-level time alignments based on the pronunciation dictionary and the orthographic transcription of each audio file. To ensure the analysis was based on reliable $f_0$ trajectories, we excluded vowel tokens for which pitch extraction failed for more than 50\% of the vowel duration (10.2\%), as well as those with durations shorter than 50 ms (5.3\%) according to established practice \citep{ting2025crosslinguistic}. Vowel segment boundaries obtained from MFA were retained as-is, unless obvious alignment errors, such as sudden changes in $f_0$ slope, were detected during manual inspection. For quality control, any utterances with mismatched transcripts, corrupted audio, or segmentation issues were removed prior to feature extraction.

\subsection{Acoustic extraction}
Audio samples were aligned by MFA into a TextGrid in Praat \citep{boersma2007praat}. For Experiment 2, which involves multiple speakers with substantially different pitch ranges, the $f_0$ data were extracted using the speaker-adapted algorithm of the Polyglot and Speech Corpus Tools \citep{mcauliffe17b_interspeech}. This algorithm involves two passes through Praat to calculate by-speaker pitch ranges for $f_0$ and then re-extract pitch using these ranges. To further ensure accuracy, we manually checked selected files from each speaker.

For Experiment 1, which compares natural speech from the LJ Speech corpus with synthetic speech generated to match the same target speaker, we intentionally did not apply speaker-specific pitch adaptation. Because both the natural and synthetic speech in this experiment originate from the same speaker, introducing different pitch parameterizations would introduce unnecessary methodological asymmetries and reduce transparency.

The analysis focused specifically on the vocalic segments of the speech signal, from which the $f_0$ values were time-normalized at 21 equidistant time points across the vowel duration\footnote{Following a widely adopted procedure in time-normalized speech analysis \citep[e.g.,][]{williams2014cross, yang2025onset}}. This sampling resolution was selected to ensure sufficient temporal granularity for capturing fine-grained phonetic variation. By normalizing $f_{0}$ trajectories in time rather than using fixed frame indices, we ensure that tokens with different vowel durations can be meaningfully compared on the same relative temporal scale. To facilitate this analysis, we selected a consistent set of vowel and onset segments across systems and frequency conditions. Table \ref{tab:segment_types} lists the specific vowels and obstruents included.

\begin{table}[h]
\centering
\begin{tabular}{l l}
\hline
\textbf{Segment Type} & \textbf{Selected Phones} \\
\hline
[+voi] obstruent &
[\textipa{b}], [\textipa{d}], [\textipa{g}], [\textipa{dZ}], [\textipa{Z}], [\textipa{v}], [\textipa{D}], [\textipa{z}] \\
$[-$voi$]$ obstruent &
[\textipa{p}], [\textipa{t}], [\textipa{k}], [\textipa{tS}], [\textipa{f}], [\textipa{T}], [\textipa{s}], [\textipa{S}], 
[\textipa{h}], [\textipa{p\super h}], [\textipa{t\super h}], [\textipa{k\super h}], [\textipa{c}], [\textipa{\c{c}}] \\
sonorant &
[\textipa{m}], [\textipa{n}], [\textipa{N}], [\textltailn], [\textipa{l}], [\textipa{j}], [\textipa{w}], 
[\textipa{m\super j}], [\s{n}] \\
vowel &
[\textipa{i}], [\textipa{i:}], [\textipa{I}], [\textepsilon], [\textipa{\ae}], [\textipa{A}], [\textipa{A:}], [\textipa{@}], [\textipa{6}], [\textipa{u}], [\textipa{U}], [\textipa{O}]

\\
\hline
\end{tabular}
\caption{IPA phones used in the analysis, grouped by segment type. This phone inventory is based on the MFA English (US) dictionary \citep{mfa_english_us_mfa_dictionary_2022}.}
\label{tab:segment_types}
\end{table}

\subsection{Statistical modeling}
We used AR1 GAMMs (generalized additive mixed models; \cite{wood2017generalized}; see \cite{soskuthy2017generalised} for introductions) to model the time-varying, potentially non-linear effects of onset voicing on the $f_{0}$ trajectory. A pooled model including data from all vowels following three target onset types (voiceless obstruent, voiced obstruent, sonorant) was fit using the \texttt{bam()} function (fREML method) in \textit{mgcv} (version 1.9.1) in R (R version 4.4.0).
Each model included a group-wise smooth spline term (by onset type, k=5)\footnote{Smooth terms were initially specified with a basis dimension of $k = 5$, following common practice in modeling normalized $f_0$ trajectories in speech. This choice generally offers sufficient flexibility to capture typical contour shapes (e.g., rising, falling, or single-peaked patterns) without overfitting. We used \texttt{gam.check()} to assess the adequacy of smoothing. The diagnostics showed no evidence of under-smoothing (all $k$-index values close to 1, all $p$-values $>$ 0.05), indicating that the specified basis dimensions were appropriate.} for onset type as predictor variables, using thin plate regression splines to capture potentially non-linear effects of onset type on $f_0$ trajectories over time.

In addition, a factor smooth (m=1, k=5) for words\footnote{Smooths were parametrized with a nonlinear penalty of order 1 (m = 1) to avoid undersmoothing by speaker differences, as suggested by \cite{wieling2018analyzing} and \cite{ting2025crosslinguistic}. Also note that the factor smooth (\texttt{bs="fs"}) for word is the setting for high-frequency group, for low-frequency group, we directly model the word effect using \texttt{bs="re"} as most words in that category have only one or
two tokens.} and vowel height (m=1, k=5) were included in the overall model as control predictors, as we expect word-specific phonetics and vowel height to have effects on the $f_0$ realization \citep{gahl2008time, jacewicz2015intrinsic}.
Difference plots for each GAMM (excluding random effects) were generated as a means of visually demonstrating where the $f_{0}$ trajectories for different onset types reached a difference significantly greater than zero (based on the default 95\% confidence level).

We divided our word list that contains 14,387 words in half based on their lexical frequency, and the lexical frequency information used to rank the word frequency was obtained from the SUBTLEX-US frequency list \citep{brysbaert2009moving}. For each onset category in each speech source, we randomly selected 1,000 tokens (detailed in Table \ref{onset:statistics}). The rationale for this approach is 1) we want to have balanced material (token) for each category as having balanced designs facilitate accurate estimation of fixed effects, reduce bias in model fitting, and ensure that smooth terms in GAMMs are estimated with adequate support. 2) to control for lexical frequency, since high-frequency words are more likely to have appeared in the training data, allowing the model to rely on surface familiarity. Low-frequency words, by contrast, test whether the model can abstract and apply prosodic patterns to unfamiliar items. 3) Incorporating all the words in a corpus as a factor-smooth term is computationally intractable and unnecessary. Each level of the factor requires estimating a separate smooth, which results in high memory consumption and increased model complexity.

\begin{table}[h]
\centering
\normalsize
\begin{tabular}{l l l l l}
\hline
\textbf{Speech source} & \textbf{Frequency} & \textbf{[+voi] obstruents} & \textbf{[-voi] obstruents} & \textbf{sonorant} \\
\hline
LJ speech       & High & 1,000 & 1,000 & 1,000 \\
FastSpeech 2    & High & 1,000 & 1,000 & 1,000 \\
Tacotron 2      & High & 1,000 & 1,000 & 1,000 \\
LJ speech       & Low  & 1,000 & 1,000 & 1,000 \\
FastSpeech 2    & Low  & 1,000 & 1,000 & 1,000 \\
Tacotron 2      & Low  & 1,000 & 1,000 & 1,000 \\
\hline
\textbf{Total}  &       & \textbf{6,000} & \textbf{6,000} & \textbf{6,000} \\
\hline
\end{tabular}
\caption{Token count per onset type across speech source and frequency bands.}
\label{onset:statistics}
\end{table}

\subsection{Additional settings for Experiment 2}
While the analysis in Experiment 1 focuses on a single-speaker training corpus, the proposed probing framework can be extended to large-scale and multi-speaker datasets. In this study, we further leverage the ‘In-the-Wild’ dataset \citep{muller2022does}, which contains both bona-fide (real) and deepfake (synthetic) audio from 58 public figures. The real-fake contrast in this dataset provides a natural baseline: bona-fide audio reveals authentic segmental $f_{0}$ perturbation patterns, while deepfake audio serves as a diagnostic probe to evaluate how well synthetic systems approximate these fine-grained phonetic effects.

In this setting, we first normalized $f_0$ values by applying z-scoring within each speaker. Then, we introduced an additional factor smooth for speaker (m=1, k=5) in the overall GAMMs. This two-step procedure enables meaningful comparison of $f_0$ trajectories across speakers, particularly in a multi-speaker setting.

\subsection{GAMM Structure and Implementation}
\begin{verbatim}
bam(zF0 ~ onset_type +   
  s(time, k=5) +
  s(time, by = onset_type, k = 5) + 
  s(time, speaker, bs=‘fs’, m=1, k=5) + 
  s(time, speaker, by = onset_type, bs = 'fs', m = 1, k=5) + 
  s(time, vowel height, bs = 'fs', m = 1, k=5) +   
  s(time, word, bs = 'fs', m = 1, k=5) +
  s(consonant, bs = 're'),
  data = data, 
  method = 'fREML', 
  rho = autocorr_m1, 
  AR.start = data$start.event)
\end{verbatim}

Above is the model structure we used for modeling onset-induced $f_0$ perturbation for the multi-speaker dataset (Experiment 2). For the single-speaker analysis in Experiment 1, we use a simpler model without by speaker factor smoothing. Autocorrelation within $f_0$ trajectories was modeled using an AR 1 structure. The autocorrelation parameter $\rho$ was estimated from a preliminary model using the \texttt{acf\_resid()} function from the \texttt{itsadug} package \citep{van2015itsadug}, and applied in the final model to account for temporal dependence in residuals.

\section{Corpus analysis}
In our experimental design, we used lexical frequency as a proxy for corpus familiarity. High-frequency words are statistically more likely to have occurred in the LJ Speech training corpus, while low-frequency words, randomly sampled from the COCA corpus, are less likely to have been encountered during training. Although this frequency-based comparison does not establish a direct causal relationship, it provides a useful basis for testing generalization capacity in the absence of explicit seen and unseen labels. In other words, we aim to evaluate whether models can generalize phonetic detail when the training data is unknown.

However, it is still important to confirm the validity of this frequency-based assumption, we performed a corpus-level analysis comparing the word list used in our study against the LJ Speech training transcripts. We found 14,127 unique words in the training corpus, and the majority of the high-frequency items were indeed found in the training corpus (70.69\% overlap), whereas most of the low-frequency items were absent (31.74\% overlap). This supports our hypothesis that lexical frequency can serve as a reasonable stand-in for training exposure. By examining how $f_0$ perturbation patterns differ between these two lexical strata, we are able to assess whether TTS systems rely primarily on memorized surface forms or if they internalize segmental-prosodic relations that support generalization to unfamiliar material. To further testify our memory-versus-generalization hypothesis, we conducted a direct comparison between words that were explicitly seen versus unseen in the training corpus. The results of this analysis are reported in fig.\ref{fig:seen_unseen} below.

\section{Analysis and results (Experiment 1)}
\label{sec:Exp1}
\begin{figure*}[t]
  \centering
  \includegraphics[width=\linewidth]{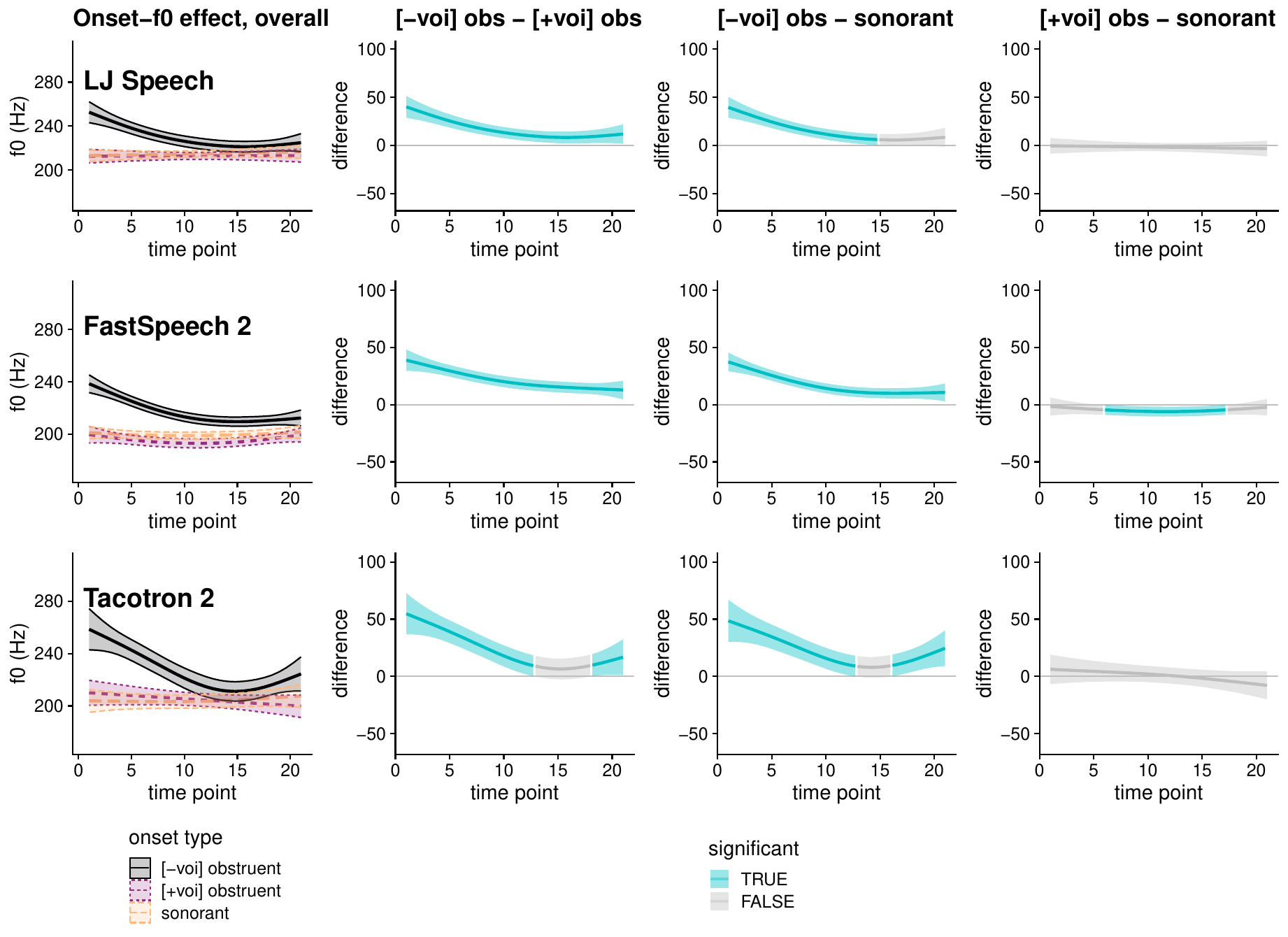}
  \caption{GAMM smooths of $f_{0}$ trajectories for onset types (leftmost column) and pairwise difference smooths (right three columns) for all \emph{high-frequency} speech sources. Shaded areas indicate significant intervals.}
  \label{fig:high}
\end{figure*}

\begin{figure*}[t]
  \centering
  \includegraphics[width=\linewidth]{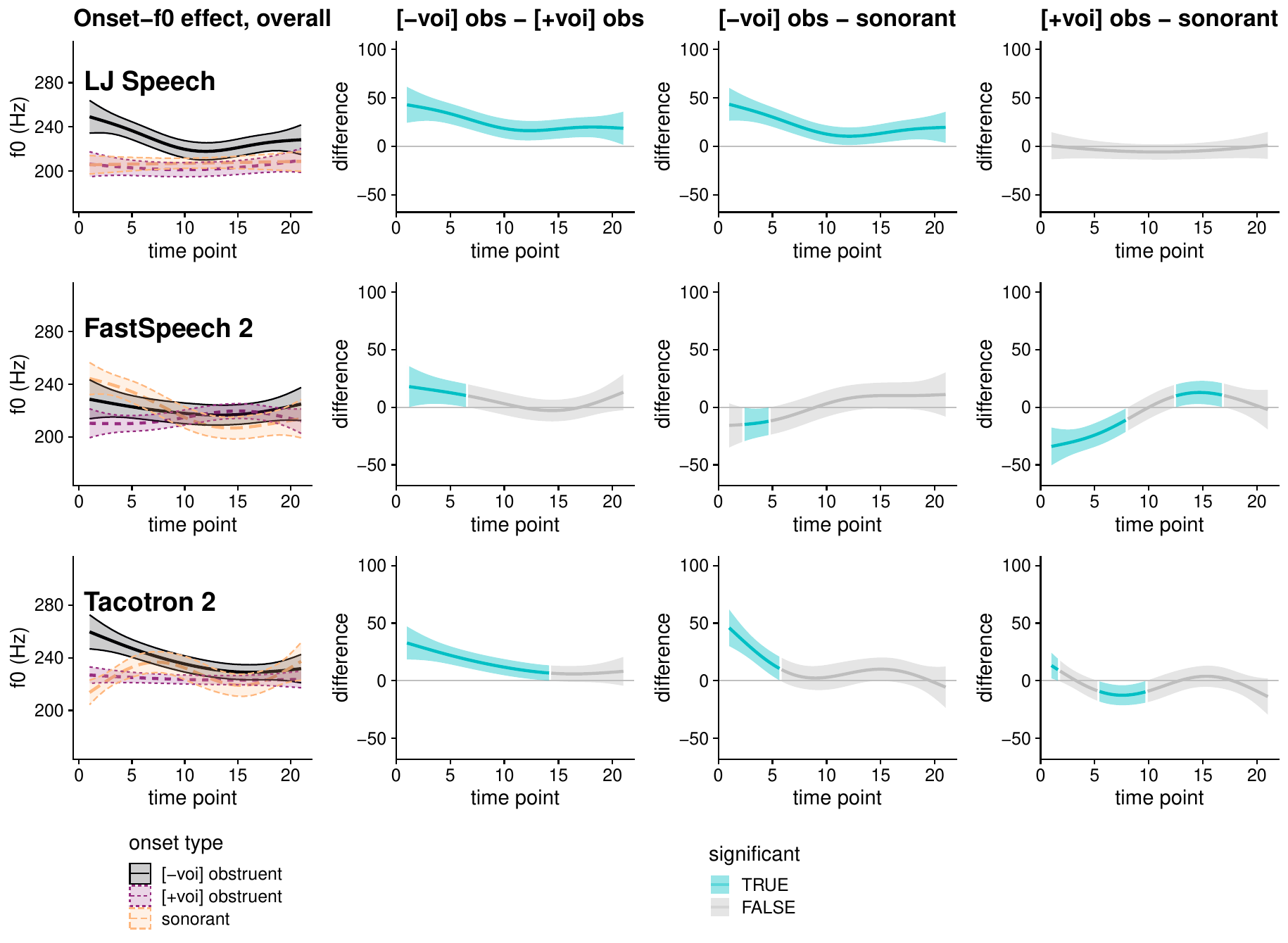}
  \caption{GAMM smooths of $f_{0}$ trajectories for onset types (leftmost column) and pairwise difference smooths (right three columns) for all \emph{low-frequency} speech sources. Shaded areas indicate significant intervals.}
  \label{fig:low}
\end{figure*}

Figures \ref{fig:high} and \ref{fig:low} illustrate the $f_{0}$ perturbation patterns for high- and low-frequency words, respectively. In these plots, the high-frequency group consists of words that are highly likely to have appeared in the LJ Speech training corpus, whereas the low-frequency group comprises words that were likely absent from the training data. The low-frequency set thus serves as a diagnostic probe for assessing whether TTS models can generalize $f_{0}$ perturbation patterns beyond familiar lexical items.

In the high-frequency condition, both the natural recordings (LJ Speech) and the synthetic outputs produced by the TTS models (FastSpeech 2 and Tacotron 2) exhibit similar $f_{0}$ perturbation patterns. Specifically, vowels following voiceless obstruents show a sharp initial $f_{0}$ peak, while those following voiced obstruents and sonorants exhibit relatively stable $f_{0}$ contours. These patterns are consistent with well-attested phonetic findings in English and other languages \citep{hanson2009effects, kirby2016effects, yang2025onset}. Additionally, the $f_{0}$ ranges across the three onset types are comparable: 190–220 Hz for voiced obstruents and sonorants, and 240–270 Hz for voiceless obstruents. These observations indicate that both models are capable of reproducing onset-conditioned $f_{0}$ patterns when the lexical items are frequent in the training data.

In contrast, for low-frequency words, the natural speech recordings maintain the expected $f_{0}$ perturbation patterns regardless of lexical frequency. However, the synthetic speech diverges considerably. FastSpeech 2 fails to produce any systematic difference in $f_{0}$ contours or height across the three onset types, suggesting a lack of learned segmental-prosodic distinctions in these unseen items. Tacotron 2 shows partial success. Although it appears to capture the elevated $f_{0}$ following voiceless obstruents, its $f_{0}$ output for vowels following sonorants and voiced obstruents is inaccurate in either overall contour shape or pitch range.

\subsection{Direct comparison of the training exposure}
As discussed above, lexical frequency provides a useful model-agnostic proxy for training exposure, particularly in realistic settings where the exact training data and procedures of TTS systems are not publicly available. However, we acknowledge that frequency-based operationalizations may introduce noise, as lexical frequency does not always perfectly align with ground-truth training-set membership (seen vs. unseen).

To further validate our interpretation and to address this potential source of noise, we additionally conduct a direct comparison based on this ground-truth exposure (seen vs. unseen). Specifically, this experiment tests the effect of training exposure by comparing tokens that are present in the LJ Speech training corpus (“seen”) with those that are absent from it (“unseen”). This analysis allows us to verify that the patterns observed using lexical frequency are not artifacts of the proxy, but reflect genuine differences in model generalization.

\begin{figure}[t]
  \centering
  \includegraphics[width=\linewidth]{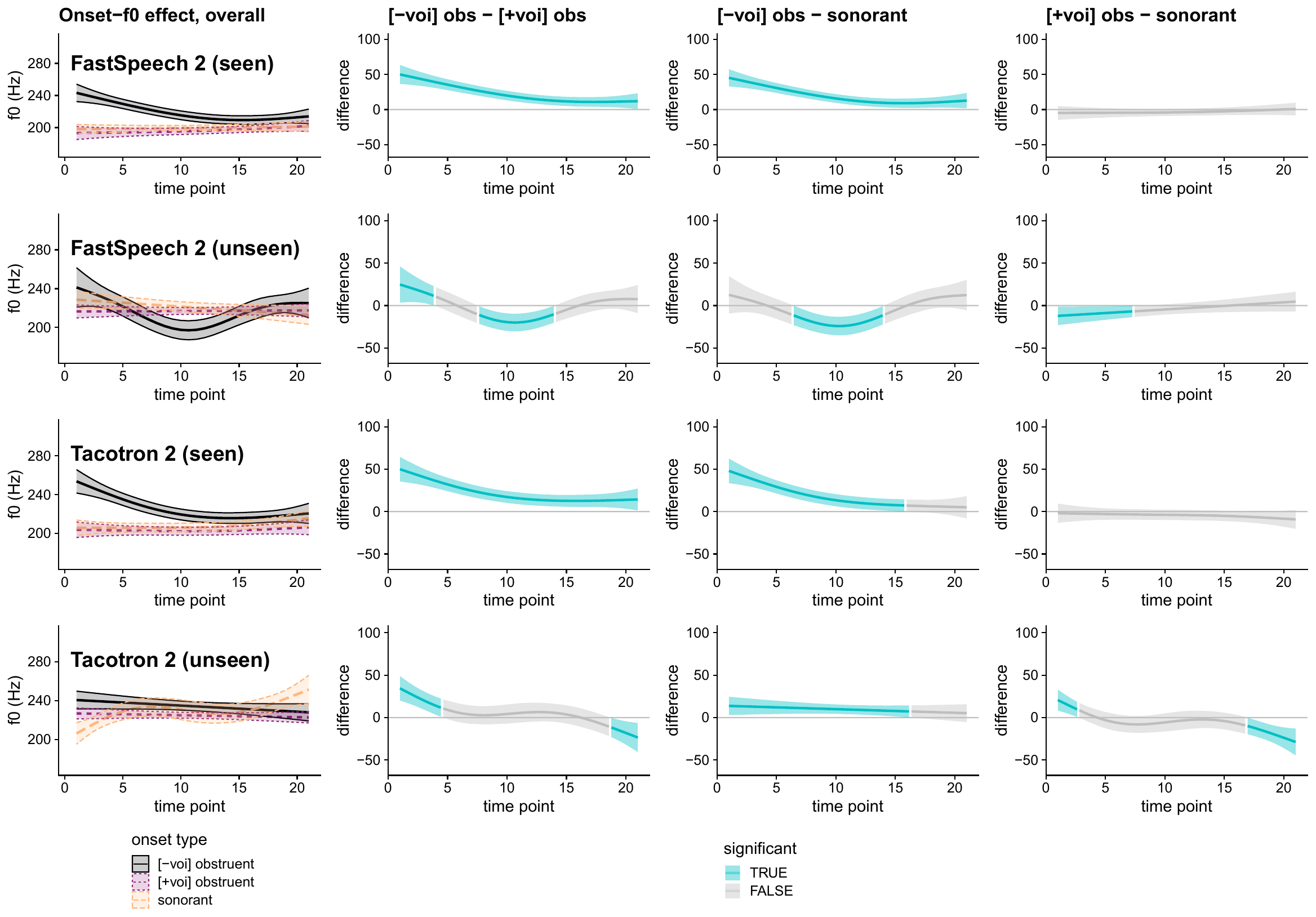}
  \caption{GAMM smooths and difference smooths for $z$-scored $f_{0}$ trajectories (seen vs unseen).}
  \label{fig:seen_unseen}
\end{figure}

The same GAMMs specification used in the main analysis was applied here to examine whether onset-conditioned $f_{0}$ patterns generalize to completely novel lexical items. Figure~\ref{fig:seen_unseen} presents the resulting smooths and difference smooths for the two conditions.

Based on the results, we found that tokens present in the training corpus (“seen”) show robust onset-conditioned separation. Vowels following voiceless obstruents begin at higher $f_{0}$ and decrease gradually, with clear and stable temporal differences relative to sonorants and voiced onsets. In contrast, tokens absent from the training corpus (“unseen”) in both FastSpeech 2 and Tacotron 2 exhibit little to no separability, with difference smooths remaining near zero across the vowel. This pattern is consistent with the frequency-stratified analysis and supports a memory-over-abstraction account of how TTS systems realize segmental–prosodic detail.

\section{Analysis and results (Experiment 2)}
\label{sec:generalize}
\begin{figure}[t]
  \centering
  \includegraphics[width=\linewidth]{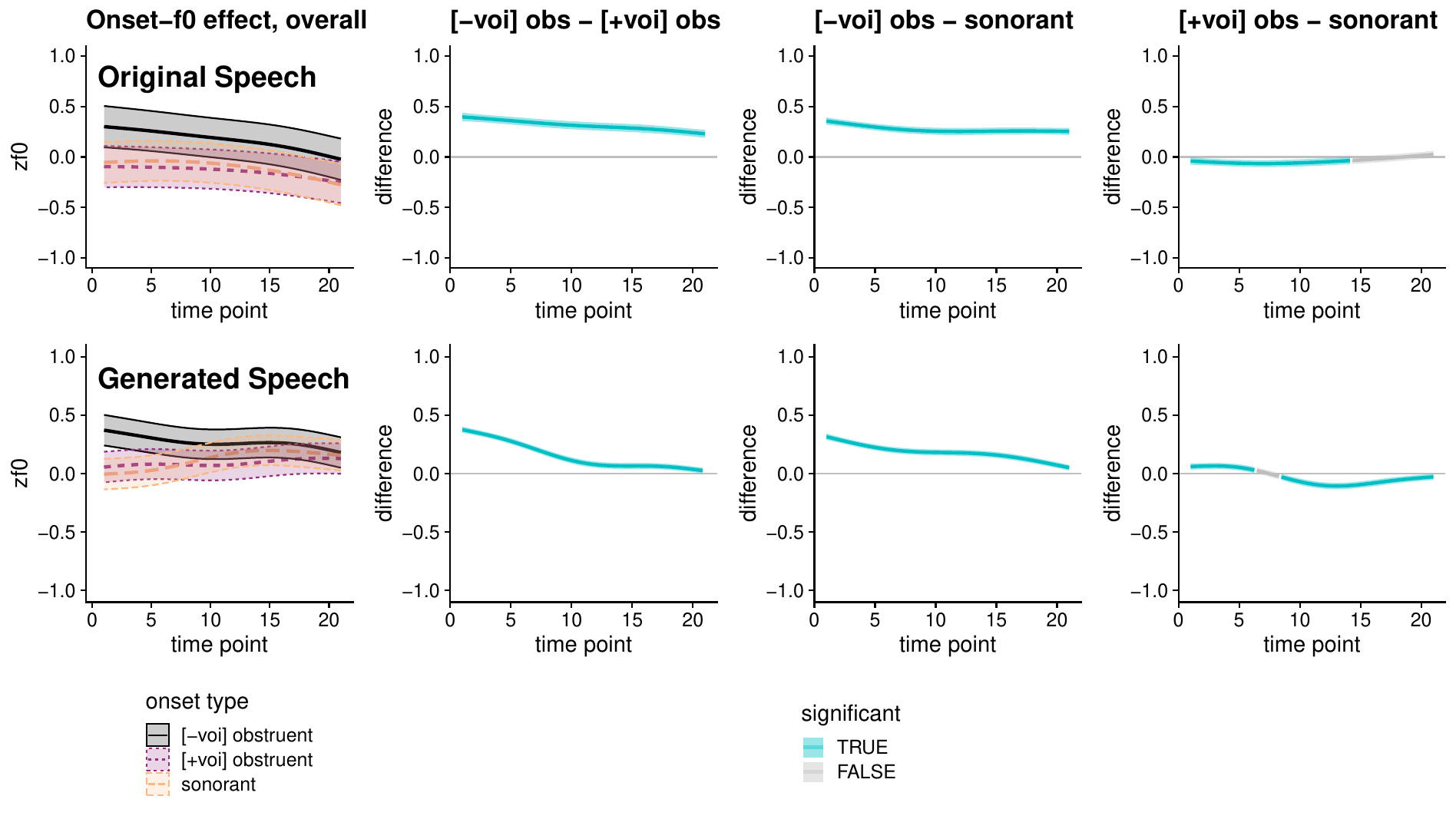}
  \caption{GAMM smooths and difference smooths for $z$-scored $f_{0}$ trajectories (high frequency group).}
  \label{fig:highinthewild}
\end{figure}

\begin{figure}[t]
  \centering
  \includegraphics[width=\linewidth]{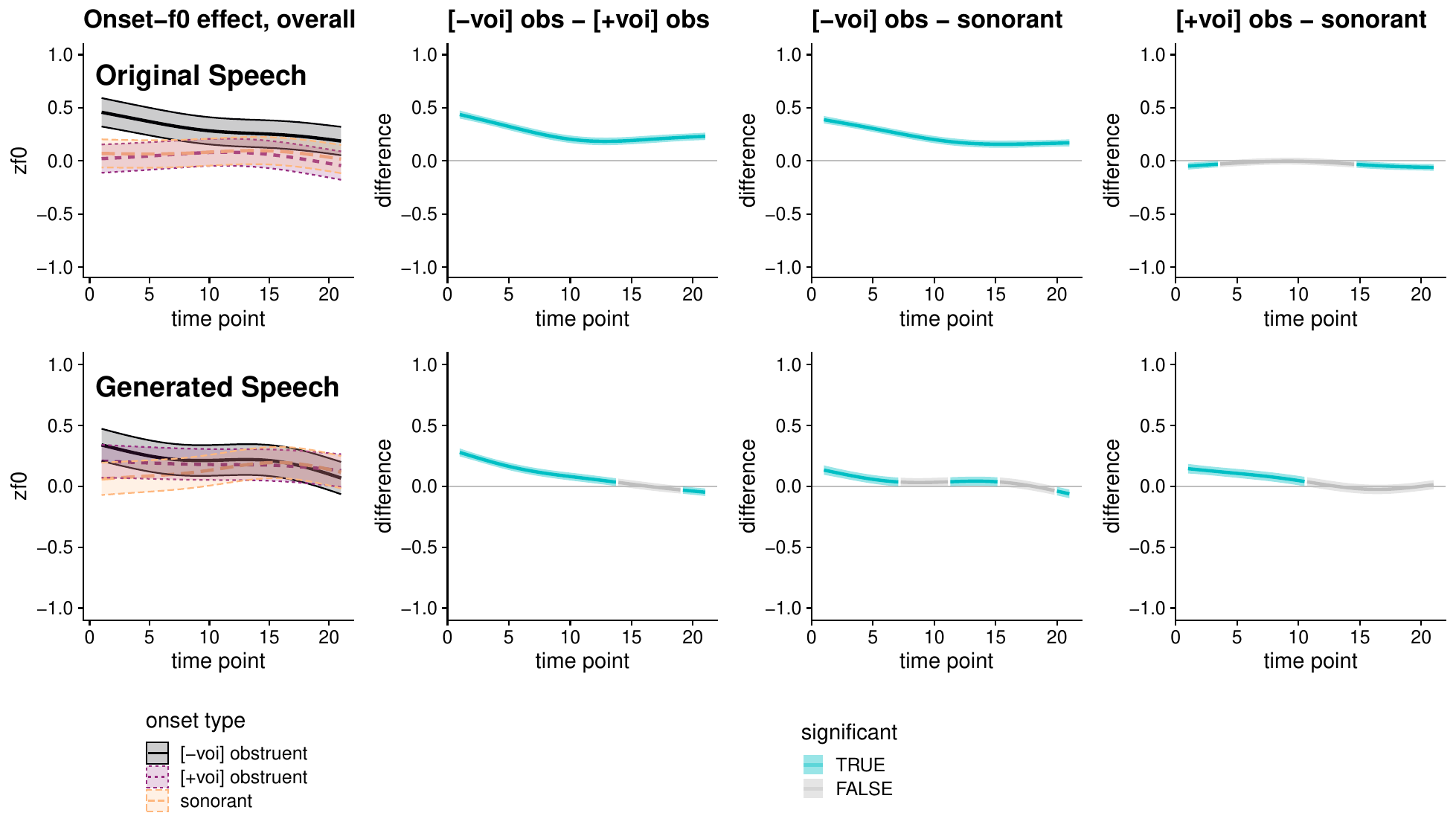}
  \caption{GAMM smooths and difference smooths for $z$-scored $f_{0}$ trajectories (low frequency group).}
  \label{fig:lowinthewild}
\end{figure}

This large-scale comparison of onset $f_{0}$ perturbation patterns is illustrated in Figures~\ref{fig:highinthewild} and~\ref{fig:lowinthewild}. In the high-frequency group (Fig. \ref{fig:highinthewild}), both the original and synthetic speech show relatively consistent $f_{0}$ trajectories across onset types. Voiceless obstruents are followed by a clear rise in $f_{0}$, while sonorants and voiced obstruents exhibit flatter or lower contours. These patterns align with the expectations established from the single-speaker corpus, suggesting that the models can successfully reproduce segmental-prosodic effects when the lexical items are likely familiar. In contrast, the low-frequency group (Fig. \ref{fig:lowinthewild}) presents a different picture. While the original speech continues to display distinct $f_{0}$ differences between onset categories, the synthetic speech shows much weaker separation. The trajectories for different onsets become more similar, and the expected rise in $f_{0}$ following voiceless obstruents is less consistent. This frequency-based contrast indicates that the ability of TTS models to generate fine-grained prosodic variation diminishes when lexical items are less familiar or unseen during training.

While the overall onset-conditioned $f_0$ patterns remain observable in the In-the-Wild dataset, the effects appear less obvious than those found in the single-speaker LJ Speech corpus. This attenuation is likely due to the large inter-speaker variability inherent in this multi-speaker dataset. In particular, the In-the-Wild corpus includes speakers from diverse linguistic and dialectal backgrounds, and the speech samples cover both formal and informal contexts. These differences can introduce prosodic variation that is not directly related to segmental conditioning, such as differences in intonational phrasing, speaking style, or register. As a result, even with speaker-level modeling, the segmental $f_0$ effects are slightly less prominent at the group level compared to results in Experiment 1. Another possible contributing factor is that the generative models used for producing speech in the In-the-Wild dataset may have been trained on substantially larger corpora, increasing the likelihood that many test words were already encountered during training. This greater overlap between training and test material could enable the models to reproduce onset-conditioned patterns more consistently from memory, thereby reducing the contrast between high- and low-frequency items and making the overall group-level pattern appear less pronounced.

These observations from the In-the-Wild dataset underscore a broader issue that extends beyond the specific systems and datasets examined here. Although this study has examined Tacotron 2, FastSpeech 2, and real versus fake (or natural versus synthetic) speech in an external deepfake dataset, the intention is not to comprehensively benchmark all available TTS models. Instead, our objective is to illustrate a broader theoretical point. When a model lacks mechanisms for encoding segmental–prosodic relationships in an explicit and structured way, it tends to fail at reproducing fine-grained phonetic effects such as consonant-induced $f_{0}$ perturbation. The absence of these patterns in low-frequency or unseen items, as observed in our experiments, suggests that this limitation is not specific to any one architecture. Rather, it reflects a tendency in the TTS systems examined here to lack sufficient structural guidance for capturing fine-grained articulatory–acoustic interactions. By treating $f_{0}$ perturbation as a diagnostic phenomenon, we offer a linguistically informed framework that can be used to evaluate phonetic generalization in future models, regardless of their underlying architecture.

\section{Discussion}
Our findings suggest that the examined neural TTS systems exhibit only limited ability to generalize segmental prosodic patterns beyond familiar lexical items. While both Tacotron 2 and FastSpeech 2 reproduce $f_{0}$ perturbation effects for high-frequency words, their performance degrades sharply on low-frequency words. FastSpeech 2 fails to distinguish between onset types altogether, and Tacotron 2 shows incomplete and unstable patterns. This contrast points to a reliance on surface-form memorization rather than abstract encoding of articulatory and acoustic interactions, raising concerns about the phonetic depth of learned representations in existing TTS architectures. The comparison between Tacotron 2 and FastSpeech 2 also illustrates that architectural differences do affect how segmental–prosodic patterns are realized, though neither model succeeds fully. The slight difference between the generation results of the two models might be due to the potential role of explicit duration and pitch predictors in FastSpeech 2, which might smooth or oversimplify local prosodic dynamics. These results suggest that neither sequential conditioning nor parallel decoding, as currently implemented, is sufficient for capturing fine-grained phonetic interactions without explicit structural guidance.

From our results, an additional natural question that arises is why, despite $f_{0}$ perturbation being a local phonetic effect and neural models being effective at modeling local acoustic correlations, neural TTS systems fail to realize this effect robustly for low-frequency or unseen words. First, we agree that, from a phonetic perspective, consonant-induced $f_{0}$ perturbation is a relatively local phenomenon. We also agree that, in principle, some signal related to this phenomenon should be present even for low-frequency items. Our interpretation, however, concerns how this knowledge is represented and under what conditions it can be reliably accessed. The results suggest that segmental-prosodic effects such as $f_{0}$ perturbation are largely encoded in a word-specific manner, tied to lexical items that are well represented in the training data. For high-frequency words, repeated exposure allows the model to reproduce stable and separable onset-conditioned $f_{0}$ patterns. In contrast, when the model encounters low-frequency or effectively nonce words, it appears unable to consistently invoke a word-independent, compositional segmental–prosodic mapping based solely on onset category.

When the model encounters low-frequency or effectively unseen words, these items function as nonce words from the model’s perspective, regardless of whether they are well-formed or even common in the human lexicon. That is, the model lacks the previously lexicalized representations (memorized words) associated with these forms and must rely solely on whatever abstract mechanisms it has learned to relate segmental context to acoustic realization.

Based on our understanding, when a TTS model encounters a nonce word that it needs to pronounce, the most important question it has to resolve is not how to model consonant-induced $f_{0}$ perturbation. Rather, the primary decision concerns where stress should be placed (on which vowel or vowels)? This is fundamentally a lexical-level problem, not a problem of fine-grained segmental interaction. Because stress assignment largely determines the overall pitch contour, the model’s priority is to ensure global prosodic coherence. As a result, to avoid unstable or erratic pitch behavior, the model might choose to neglect some explicit segmental-level interaction mechanisms. In other words, suppressing consonant-specific $f_{0}$ perturbation could be a modeling decision of TTS systems.

To illustrate the contrast, imagining pronounce a nonce word such as blicket or splone. Although these words are novel, human speakers can pronounce them in a highly systematic way. For example, their realization still obeys well-established phonetic regularities: articulatory constraints on vocal fold tension, aerodynamic requirements for voicing, and other biomechanical properties of the speech which ensure that consonant-induced $f_{0}$ perturbation, among other effects, emerges naturally even in completely novel lexical items. In addition, psychological and phonological constraints allow speakers to generalize learned sound patterns and apply them productively to new forms.

Current neural TTS systems, however, lack access to these articulatory and cognitive constraints. They do not possess an explicit representation of vocal fold dynamics, nor do they encode phonetic rules as abstract, compositional mappings that can be flexibly applied to unseen inputs. As a result, when confronted with a nonce word, the model does not “reason” about how segments should interact based on articulatory principles. Instead, it defaults to an overall statistical tendency shaped by its training data, producing a plausible pronunciation at a global level but failing to consistently instantiate fine-grained segmental–prosodic relations such as onset-conditioned $f_{0}$ perturbation.

From this perspective, the contrast between human and model behavior is not that models cannot pronounce nonce words at all, but that they do so via fundamentally different mechanisms. Human speakers generate novel words by deploying abstract phonetic knowledge grounded in physiological and cognitive constraints, whereas current TTS models rely on lexicalized and distributional patterns that do not generalize reliably once lexical support is removed. Therefore, the observed frequency effect does not contradict the ability of neural models to learn local correlations. Instead, it highlights a limitation in the generalization and accessibility of such correlations across the lexical space. While weak or implicit CF0-related signals may still be present for low-frequency items, they do not surface as stable, robust, and statistically separable output patterns in group-level modeling.

Such limitations can be addressed either by substantially increasing the diversity and quantity of training data, or by introducing explicit mechanisms for modeling segmental–prosodic interactions. The former may allow models to learn generalizable patterns through exposure alone, while the latter offers a more principled approach by guiding the model to encode phonologically meaningful dependencies that are otherwise underrepresented in the data.

The observed failure of TTS models to generalize $f_{0}$ perturbation to low-frequency words has broader implications beyond phonetic evaluation. First, it highlights a potential vulnerability of neural TTS systems to detection: the absence or degradation of segmental-prosodic effects in rare or novel words may serve as a reliable acoustic cue for distinguishing synthetic from natural speech. This offers a promising direction for developing segmental-level detection tools that go beyond global prosodic or spectral statistics and target fine-grained articulatory correlates. Such segment-based detection is both linguistically grounded and difficult for the current TTS model to accurately reproduce or strategically suppress \citep{YANG2026112768}.

Second, the findings raise the possibility of reverse engineering model characteristics based on acoustic output. For example, the presence or absence of specific phonetic detail, such as onset-driven $f_{0}$ modulation, may allow external observers to infer properties of the underlying model, including its architecture, the extent of exposure to particular lexical items, or the granularity of prosodic supervision to some extend. Similarly, the asymmetry in $f_{0}$ realization across lexical frequency strata suggests that the expressive capacity of a TTS model may be constrained by the lexical and phonetic diversity of its training data. In this sense, systematic probing of segmental effects could be used not only to assess generalization, but also to estimate the effective size or coverage of the training corpus. Similar approaches have been used in other domains. For instance, auditing text-generation models by querying whether a user’s texts were part of the training data \citep{song2019auditing}, or using membership inference attacks to estimate training data memorization in masked language models \citep{mireshghallah2022quantifying}. This opens the door to using phonetic probes as a lens for analyzing TTS training data indirectly, which may have implications for data transparency, privacy auditing, and model interpretability.

Finally, an important direction for future work could be the perceptual consequences of failing to reproduce fine-grained segmental effects such as consonant-induced $f_{0}$ perturbation in synthetic speech. Although TTS quality as measured by mean opinion scores (MOS) has improved greatly with the advent of modern neural architectures, a persistent gap remains relative to natural speech, particularly for small-scale models such as those examined in this study. While MOS captures overall naturalness and listener preference, it is unspecific to the specific acoustic cues that contribute to these global impressions.

Previous work in prosody modeling and expressive TTS has demonstrated that richer and more dynamic pitch contours are associated with higher subjective naturalness and expressiveness \citep{huang2023prosody}. In the broader speech perception literature, dynamic $f_{0}$ variation is known to influence intelligibility and listener preference, with flattened or unnatural pitch contours leading to reduced perceived quality in adverse listening conditions \citep{laures2003perceptual}. These findings align with the intuitive view that sublexical pitch patterns contribute to naturalness judgments, even if they are subtle and not directly captured by aggregate metrics such as MOS. On this basis, it is plausible that the inability of neural TTS systems to consistently generalize local phonetic phenomena to low-frequency or unseen items may contribute in part to the overall MOS gap between synthetic and natural speech. Establishing a direct causal link between segmental-level deviations and listener judgments would require targeted perceptual experiments, such as controlled listening tests or systematic parameter manipulations.

Future research could also explore architectural biases, targeted training objectives, and pretraining strategies to enhance the phonetic interpretability and generalization of neural TTS systems. In addition, future research could extend the present experiment to other TTS architectures under controlled training settings, such as diffusion-based and flow-based generative models, to assess whether the observed patterns generalize beyond the models examined here.

\section{Conclusion}
This study examined whether neural TTS systems are capable of reproducing segmental-level $f_{0}$ perturbation, a phonetic effect arising from local articulatory mechanisms. By comparing Tacotron 2 and FastSpeech 2 against natural speech, we found that both models can replicate onset-driven $f_{0}$ patterns in high-frequency words, but fail to generalize these effects to low-frequency items. These findings suggest that TTS systems rely more on lexical memorization rather than abstract phonetic encoding. Moreover, neither autoregressive nor non-autoregressive architectures, as currently implemented, are sufficient for learning fine-grained segmental–prosodic interactions without explicit guidance. Our proposed probing framework offers a linguistically grounded diagnostic tool for evaluating such interactions, with potential applications in TTS detection, model auditing, and corpus analysis.

\appendix
\section{Data availability and software setup}
\label{app1}

All the speech corpus, lexical corpus, and speech models used in this study are publicly available online, and we have provided corresponding references in the experimental setting section for the purpose of reproduction.

\section{Acknowledgment}
We would like to thank the anonymous reviewers for their constructive comments and helpful suggestions. We are also grateful to Matthew Faytak for helpful comments and discussions that later proved valuable in shaping some of the ideas developed in this paper.

\newpage
\bibliographystyle{elsarticle-harv}
\bibliography{references}

\end{document}